\documentclass[10pt,journal,compsoc]{IEEEtran}

\ifCLASSOPTIONcompsoc
  \usepackage[nocompress]{cite}
\else
  \usepackage{cite}
\fi

\usepackage[utf8]{inputenc} 
\usepackage[T1]{fontenc}    
\usepackage{hyperref}       
\usepackage{url}            
\usepackage{booktabs}       
\usepackage{amsfonts}       
\usepackage{nicefrac}       
\usepackage{microtype}      
\usepackage{xcolor}         

\usepackage{bm}
\usepackage{amsmath}
\usepackage{multirow}
\usepackage{graphicx}
\usepackage{makecell}
\usepackage{enumitem}
\usepackage{amsthm}

\usepackage{color}

\begin{document}

\title{Discovering Predictable Latent Factors for Time Series Forecasting}

\author{Jingyi~Hou,
        Zhen~Dong,
        Jiayu~Zhou,
        and~Zhijie~Liu,~\IEEEmembership{Member,~IEEE}
\IEEEcompsocitemizethanks{\IEEEcompsocthanksitem 
J. Hou and Z. Liu are with the School of Intelligence Science and Technology, University of Science and Technology Beijing, Beijing 100083, China, with the Institute of Artificial Intelligence, University of Science and Technology Beijing, Beijing 100083, China, and with the Key Laboratory of Intelligent Bionic Unmanned Systems, Ministry of Education, University of Science and Technology Beijing, Beijing 100083, China.
E-mail: \{houjingyi, liuzhijie\}@ustb.edu.cn.\protect\\
\IEEEcompsocthanksitem Z. Dong is with the College of Mathematics and Computer Science, Yan’An University, Yan’an 716000, China.
E-mail: cndongzhen@gmail.com\protect\\
\IEEEcompsocthanksitem J. Zhou is with the Department of Computer Science and Engineering, Michigan State University, East Lansing, MI 48823. 
E-mail: jiayuz@msu.edu.
}
}

%

\IEEEtitleabstractindextext{%
\begin{abstract}
  Modern temporal modeling methods, such as Transformer and its variants, have demonstrated remarkable capabilities in handling sequential data from specific domains like language and vision.
  Though achieving high performance with large-scale data, they often have redundant or unexplainable structures.
  When encountering some real-world datasets with limited observable variables that can be affected by many unknown factors, these methods may struggle to identify meaningful patterns and dependencies inherent in data, and thus, the modeling becomes unstable and unpredictable.
  To tackle this critical issue, in this paper, we develop a novel algorithmic framework for inferring latent factors implied by the observed temporal data. 
  The inferred factors are used to form multiple predictable and independent signal components that enable not only the reconstruction of future time series for accurate prediction but also sparse relation reasoning for long-term efficiency.
  To achieve this, we introduce three characteristics, i.e., predictability, sufficiency, and identifiability, and model these characteristics of latent factors via powerful deep latent dynamics models to infer the predictable signal components.
  Empirical results on multiple real datasets show the efficiency of our method for different kinds of time series forecasting tasks.
  Statistical analyses validate the predictability and interpretability of the learned latent factors.
\end{abstract}

}

\maketitle

\IEEEdisplaynontitleabstractindextext

%
\IEEEpeerreviewmaketitle

\IEEEraisesectionheading{\section{Introduction}\label{sec:introduction}}

\IEEEPARstart{T}{ime} series forecasting is an essential task of data science and machine learning, with broad applications spanning areas like financial management, health informatics, weather forecasting, and epidemic prediction.
Deep learning has shown its strong ability to model various complex sequential information in many fields, especially for vision and language tasks \cite{DBLP:conf/iccv/Feichtenhofer0M19,DBLP:conf/emnlp/0001DL16,DBLP:conf/nips/VaswaniSPUJGKP17}. 
Even though there are many efforts applying deep learning in time series forecasting, unlike the data in successful deep learning applications, most real-world time series neither contain sufficient information with high-dimensional data nor represent explicit semantic information that can guide the data analysis. 
Therefore, deep learning methods still face critical challenges in time-series forecasting tasks.

A particular challenge lies in the inherent spatio-temporal complexity of time series data, as the underlying patterns can be influenced by a multitude of unknown factors. 
Discovering and modeling these factors and their relations empower time series forecasting with good interpretability and thus greatly enhance the forecasting performance of models.
Prior work has delved into the use of factors and their relations to forecast time series, either by explicitly exploiting additional knowledge \cite{DBLP:conf/acl/CohenX18,DBLP:journals/tkde/ShiTWZB19,DBLP:journals/pr/ChengYXL22,DBLP:journals/corr/abs-2110-13716} or explicitly modeling relations of latent representations \cite{duan2022factorvae}.
These approaches may have limited applicability and are often feasible for specific tasks with the need for a sophisticated process of domain knowledge.
However, real-world time series could be potentially affected by various kinds of factors from different domains, and it is infeasible to explicitly collect a comprehensive set of all the factors.
For example, in the financial domain, asset volatility describes the price changes of an asset (e.g., a stock), and its prediction is important in many quantitative strategies. 
Asset volatility is affected by massive factors that are hard to observe or measure, e.g., macro economy, investor sentiment, and various events in the markets.
Moreover, real-world relations between different factors over time are highly complex, with some factors being unobservable, which further complicates precise modeling.

Fortunately, since some observable data are dependent on these unobserved factors, it is possible to learn a graphical model that can infer latent factors from observations generatively.
This allows us to learn and disentangle latent factors with characteristics that support rational relation modeling to explain the generation procedure of time series, which benefits further forecasting.
Specifically, the inferred latent factors are used to form various simple and intrinsic temporal signals that are easy to forecast, where ``intrinsic'' denotes factors in the same signal component are self-descriptive and not affected by factors from other signal components.
Through this straightforward inductive bias, we can make reasonable independence assumptions on these latent factors to encourage sparsity in their relations, reducing computational complexity.
As the derived temporal signals are simple and intrinsic, it is easy to forecast the future factors of each signal, and therefore, predict future data given the predicted factors.

{
Numerous efforts have been dedicated to decomposing temporal signals into sub-components that are more steady or regular than the original signal.
Recent advancements in deep time series forecasting methodologies \cite{asadi2020spatio,DBLP:conf/iclr/OreshkinCCB20,DBLP:conf/nips/WuXWL21} employ seasonal-trend decomposition \cite{cleveland1990stl}, which partitions time series into interpretable and reliable components, including trend, seasonal, and residual elements.
However, decomposition based on predefined rules might fail in the scenario where time series are affected by various factors and barely show explicit regularities.
For example, Zeng \emph{et al.} \cite{DBLP:conf/aaai/ZengCZ023} show that for financial time series forecasting, state-of-the-art methods demonstrate suboptimal performance compared to a basic approach that simply relies on repeating the last observed value within the look-back window. 
The non-stationary signal decomposition techniques, including empirical mode decomposition (EMD) \cite{huang1998empirical}  and its variations \cite{huang2014hilbert,DBLP:journals/tsp/DragomiretskiyZ14}, effectively decompose time series into stationary intrinsic mode functions (IMFs) that dynamically represent the local oscillatory patterns at different time scales.
Neural basis expansion analysis approaches \cite{DBLP:conf/iclr/OreshkinCCB20,DBLP:journals/corr/abs-2201-12886} hierarchically decompose the input signal into easily reconstructable sub-signals with the doubly residual mechanism.
Regardless of whether deep learning techniques are employed,  these empirical decomposition methods focus on the characteristics of the original time series, while our method pays attention to drawing patterns from the observations in statistics.
Beyond signal decomposition, we endeavor to automatically infer easily predictable factors from complex time series data by modeling probabilistic relations between these factors and observations.
}
In addition, we assume pairwise independence between factors of different derived temporal signals to ensure the signals' intrinsicality and sparse relations for efficient computation.

Accordingly, we aim to discover latent factors to form multiple signal components that are simple and intrinsic for prediction (i.e., predictability), informative for reconstruction (i.e., sufficiency), and feasible for learning (i.e., identifiability). 
To achieve this, we propose a novel approach for inferring predictable factors for forecasting time series.
The proposed framework is constructed based on variational inference and sequence models by approximating the joint distribution over observed time series and latent factors following the aforementioned key characteristics of predictability, sufficiency, and identifiability.
The framework can also be decomposed into multiple latent dynamics models that are conditionally pairwise independent for alleviating the error accumulating of the long-term series.
We conduct empirical studies on two typical time series forecasting tasks, i.e., long-term series forecasting and stock trend forecasting, to show the effectiveness of our method, and we also conduct statistical analyses to gain insights into the predictability and interpretability of the inferred signal components.

The contributions of this work are summarized as follows:

\begin{itemize}[leftmargin=1.25em]
  \item We consider inferring latent factors varying along time from the observation to form multiple signal components with characteristics of predictability, sufficiency, and identifiability. The factors we inferred can be used to easily construct future signals for time series forecasting.
  \item We design a novel method that disentangles the latent factors and models the relations of the factors according to the signal components to draw the joint distribution for more efficient approximation.
  \item The proposed method achieves better performance compared with the state-of-the-art for long-term series forecasting and stock trend forecasting, respectively.
\end{itemize}

\section{Related Work}

\subsection{Time Series Forecasting}

Time series forecasting is challenging and has long been investigated in many research areas. 
Several classical methods are proposed for industries, such as the autoregressive methods \cite{sims1980macroeconomics,tauchen1986finite}, moving average methods \cite{box2015time}, filtering models \cite{morrison1977kalman,joo2015time}, and factor models \cite{sharpe1964capital,fama1992cross}.
Although these methods are limited in their ability to handle data available nowadays that become increasingly large-scale and high-dimensional, they still provide theoretical guidance for time series forecasting methods.

With the notable rise of deep learning, sequence-based neural networks are getting increasingly involved.
One typical structure is the recurrent neural network (RNN) which is able to process theoretically arbitrary long-term input sequences with only a few parameters \cite{DBLP:conf/nips/RangapuramSGSWJ18,DBLP:journals/corr/FlunkertSG17}.
Despite these advantages, the recursive mechanism is not really conducive to information propagation.
Attention-based RNNs \cite{DBLP:conf/ijcai/QinSCCJC17,DBLP:conf/aaai/SongRTS18,DBLP:conf/ijcai/FengC0DSC19} address this problem by encoding long-term series with adaptive weighted summation to generate predictions.
Another structure is the convolutional network that captures various local temporal patterns with different convolutional filters \cite{borovykh2017conditional,DBLP:conf/sigir/LaiCYL18,DBLP:conf/nips/SenYD19}.
Recently, Transformer-based models \cite{DBLP:conf/ijcai/FengC0DSC19,DBLP:conf/iclr/KitaevKL20,DBLP:conf/aaai/ZhouZPZLXZ21,DBLP:conf/nips/WuXWL21,liunon,DBLP:conf/icml/ZhouMWW0022} have emerged, treating time series as undirected graphs with nodes as time steps and edge weights as attention between nodes.
Although Transformers have strong model-fitting abilities, they still suffer from high computation complexity.
To reduce parameters, RNNs, convolutional networks, and Transformers make some assumptions (e.g., sparsity and locality) at the cost of impairing the long-term modeling capability.
Differently, our method focuses on inferring easy-to-predict factors from the original time series, so that we can use these relatively simple predictors (without strong fitting ability) to process the factors and then calculate the future data according to the predicted factors.

The structures of our methods can be regarded as deep latent dynamics models \cite{DBLP:conf/icml/HafnerLFVHLD19,DBLP:conf/icml/BeckerPGZTN19,DBLP:conf/nips/SaxenaBH21} by considering the original time series as observations and the prediction of each factorized signal component as the decision-making process.
This helps mitigate accumulating errors and improve the accuracy of long-term predictability estimation.
There are methods \cite{DBLP:conf/aistats/KhemakhemKMH20,DBLP:conf/nips/HalvaCLSZGH21} studying the non-linear independent component analysis (ICA) to disentangle latent factors of dynamic models. 
The disentanglement can be effectively learned by these models thanks to their theoretical guarantees of the models' identifiability.
Nevertheless, these methods mainly focus on designing generalized models without being constrained by too many restrictive assumptions. 
Differently, our method focuses on using the disentanglement of the factors to model reasonable and sparse relations for more efficient approximation.

Some methods use GNNs \cite{DBLP:conf/cikm/ChenWH18,DBLP:journals/corr/abs-1908-07999,DBLP:conf/kdd/WuPL0CZ20,DBLP:conf/iclr/ChenSCG22} or MRFs \cite{DBLP:conf/kdd/LiST19} to model spatiotemporal relations between different time series variables to improve the forecasting performance.
However, reasoning and propagating information through global relations could incur computational costs and introduce redundancy as data scale and dimension increase.
We propose to use variational inference to draw distributions over latent factors and learn relations between them to generate and predict time series.
Duan \emph{et al.}~\cite{duan2022factorvae} leverage VAEs to learn the multi-dynamic-factor model \cite{ng1992multi} for predicting cross-sectional stock return, ignoring the temporal variation of factors.
Differently, our method takes the temporal relations into account for more comprehensive modeling and adapts the classic theory with fewer constraints by incorporating deep learning techniques.

There are methods using extra knowledge of the market to improve the performance of times series forecasting, such as the event-driven approaches \cite{DBLP:conf/acl/CohenX18,DBLP:journals/tkde/ShiTWZB19,DBLP:journals/pr/ChengYXL22} and the method exploiting additional enterprise information \cite{DBLP:journals/corr/abs-2110-13716} for stock trend forecasting.
Although these methods utilize most information from the environment, our method has the advantage of discovering complementary factors by analyzing the data along the temporal dimension to further boost the prediction performance, as evidenced by our experiments.

\subsection{Decomposition of Time Series}

Decomposing non-stationary observed time series into simple and predictable components is quite an intuitive way for time series analysis, and can be traced back to the blind source separation \cite{jutten1991blind}.
Empirical mode decomposition methods \cite{huang1998empirical,huang2014hilbert,DBLP:journals/tsp/DragomiretskiyZ14} iteratively decompose time series into stationary oscillating components, known as intrinsic mode functions (IMFs).
The IMFs are simple and nearly orthogonal signals and are defined to have a similar number of extrema and zero crossings as well as symmetric envelopes to guarantee the conduct of the Hilbert transform on them.
Instead of the delicately predefined intrinsic modes that should be oscillatory and add up to be the original signal, with the help of deep learning techniques, our method generalizes the data processing from the frequency domain to the concept domain by preserving the simplicity and orthogonality of the intrinsic modes that are in accordance with our instincts.

Researchers have recently ventured into constructing deep networks for time series decomposition. 
In these endeavors, various approaches \cite{asadi2020spatio,DBLP:conf/iclr/OreshkinCCB20,DBLP:conf/nips/WuXWL21,DBLP:conf/icml/ZhouMWW0022} explicitly decompose time series into trend and seasonality components, subsequently utilizing these components to forecast future signals.
Sen \emph{et al.}~\cite{DBLP:conf/nips/SenYD19} use temporal convolution networks to perform matrix factorization for prediction time series.
Wang \emph{et al.}~\cite{wangmicn} use isometric and multi-scale convolutions to capture local-global interactions within time series data.
These methods mainly focus on leveraging the existing classic signal processing theories to enhance model interpretability, while our method harnesses the potent information abstraction capabilities of deep learning to extract more intrinsic representations as predictable latent factors for time series forecasting.
{
Except for the season-trend decomposition, Transformer-based methods, i.e., Autoformer \cite{DBLP:conf/nips/WuXWL21} and FEDformer \cite{DBLP:conf/icml/ZhouMWW0022},  additionally introduce new attention mechanisms as inner decomposition modules according to the frequency characteristics of signals.
}
Neural basis expansion analysis approaches methods like N-BEATS \cite{DBLP:conf/iclr/OreshkinCCB20} and N-HITS \cite{DBLP:journals/corr/abs-2201-12886}, use residual mechanisms to hierarchically decompose predictions into interpretable outputs augmenting the seasonality-trend decomposition paradigm.
Unlike gradually decomposing the original signal, our method infers latent factors to form intrinsic and predictable signal components from the input data with the guidance of semantics, which is more flexible for relation modeling.

\subsection{Multi-Scale Temporal Sampling}

{Since the multiple signal components are generated using a multi-scale sampling strategy in this paper, we review the sequence prediction methods that also benefit from this strategy.
Temporal convolutional network (TCN) \cite{DBLP:journals/corr/abs-1803-01271} stacks dilated convolutions with different dilation rates that can capture multi-scale temporal information for vision and language modeling and prediction.
TS2Vec \cite{DBLP:conf/aaai/YueWDYHTX22} uses hierarchical contrastive learning to capture scale contextual information of time series.
While these methods mainly focus on extracting local contextual patterns with different temporal granularities, our method uses multi-scale sampling to discover multiple latent semantics with different periodic properties hidden in time series.
Methods cascade multi-scale sampling in RNNs, such as DilRNN \cite{DBLP:conf/nips/ChangZHYGTCWHH17} and LSTNet \cite{DBLP:conf/sigir/LaiCYL18} to further improve long-term memory capability of models.
N-HITS \cite{DBLP:journals/corr/abs-2201-12886} applies multi-scale sampling to the input signal to construct the prediction hierarchically.
Compared to the above methods, we take full advantage of multi-scale sampling as an inductive bias to guide the learning of latent factors with the time-scale semantics and the modeling of sparse relations between latent factors.  
}

\section{Predictable Latent Factor Inference Modeling}
\label{sec:plfi}


\begin{figure*}[tbp]
\centerline{\includegraphics[width=1.65\columnwidth]{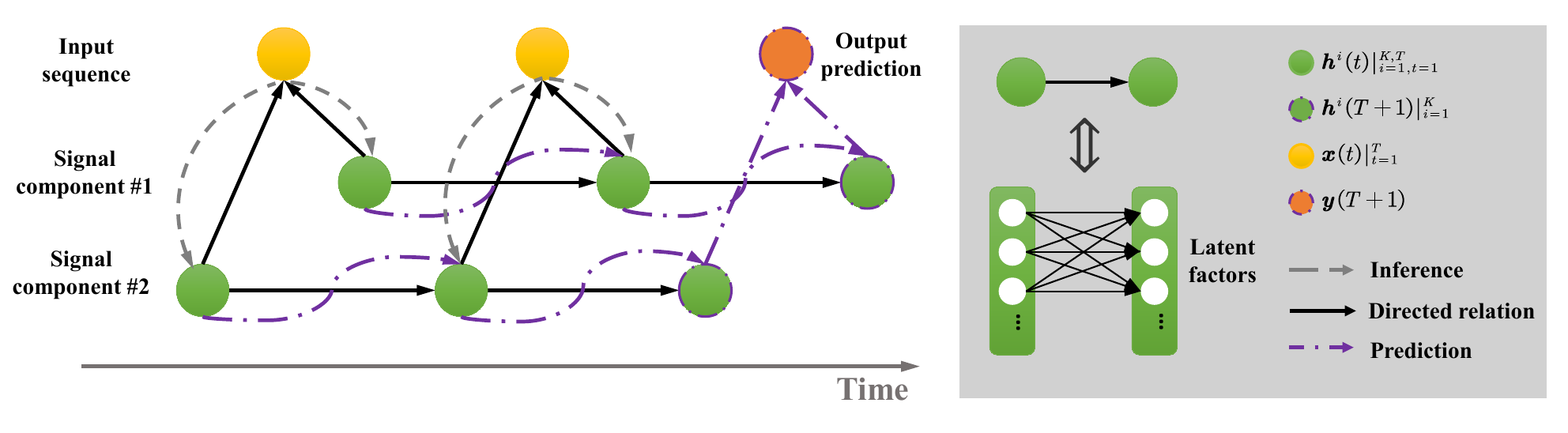}}
\caption{The graphical model for learning predictable latent factors. We present a model with 2 signal components for illustration here.}
\label{fig:framework}
\end{figure*}

We aim to discover predictable latent factors of time series data for efficient forecasting.
To achieve this, we first model the relations among the latent factors and the observed time series, and then learn the latent factors from the time series based on the modeled relations.
Specifically, we assume latent factors can be divided into several independent signal components, with conditional independence between factors within each component for computational efficiency.
Since the inferred latent factors drive the trends of the times series, the original time series data can be analyzed in a latent semantic domain. 
To effectively learn latent factors for feasible and effective prediction, we enforce the factors in each signal component with the characteristics of predictability, sufficiency, and identifiability, which will be elaborated in Section \ref{sec:PLFl}.

Fig.~\ref{fig:framework} shows the graphical model of the proposed predictable factor learning method.
Let $\bm{X} = [\bm{x}(1),\dots,\bm{x}(T)]\in \mathbb{R}^{D\times T}$ be the matrix of input time series of past $T$ time steps with $D$ representing the number of features, and $\bm{H} = [\bm{h}(1),\dots,\bm{h}(T)]\in \mathbb{R}^{L \times K \times T}$ are the factorized $K$ signal components each of which is composed of $L$ predictable latent factors.
The observed time series $\bm{X}$, e.g., the price signal of stocks, are usually non-stationary time series that cannot be predicted, and we assume that $\bm{X}$ depends on the hidden predictable $\bm{H}$ which can be inferred from $\bm{X}$ to accomplish time forecasting tasks.
With the predicted $\bm{h}(T+1)$, we can make a specific prediction $\bm{y}(T+1)$ according to different kinds of time series forecasting.

\subsection{Multi-Scale Convolutional Encoder}
Computational efficiency is among the main challenges in time series analysis. To improve the computational efficiency, we thereby make several independent assumptions to encourage the sparsity of the relations between the inferred factors.
The first independent assumption is that factors representing different time-scale semantics are merely related.
Taking the financial time series data as an example, the market sentiments and the business scopes are both causal factors of the stock prices while they are independent.
Market sentiments might have weekly seasonality, e.g., hesitant on Mondays and active on Wednesdays.
And the profits of companies with specific business scopes usually have yearly seasonality, e.g., relatively high in summer for a company selling air-conditioners.
More supportive statements are claimed by \cite{muller1993fractals}.
The second assumption we make is that factors in the same time scale are conditionally independent.
This assumption leverages temporal information to clearly describe the directed relations between latent factors.  
If two factors are related to each other, the factor at the previous time step determines the current time step, and these two factors are independent given the observations at the current time step.

With the two independence assumptions above, we now discuss the use of a multi-scale convolutional encoding network, which can discover various predictable latent factors and directly divide the factors into different independent signal components according to the time scales.
Concretely, we sample the observed time series data in several specific time scales corresponding to different semantics, such as ``day'' and ``week''.
Let $\mathcal{R}=\{r_{1},\dots,r_{K}\}$ be the sets of $K$ sampling rates, the sampled time series are $s(\bm{X}, \mathcal{R})=\{s(\bm{X},r_{i})|i=1,\dots,K\}$, where $s(\bm{X},r)$ describes the sampling operation on $\bm{X}$ with sampling rate of $r$. 
The proposed encoder $\phi$ learns latent factors at each time scale individually, and the factors are combined during the decoding procedure.
In particular, the $\phi$ is composed of $K$ sub-encoders $\phi = \{\phi^{(i)}\}|_{i=1}^{K}$, and these sub-encoders are used to learn the distributions of latent factors:
\begin{equation}
\begin{aligned}
\label{eq:encoder}
  &[\bm{\mu}^{i},\bm{\sigma}^{i}] = \phi^{(i)}(\bm{X}^{(i)}), \ \bm{X}^{(i)} = s(\bm{X},r_i),\\
  & \bm{h}^{i}_j(t) \sim \mathcal{N}(\bm{\mu}^{i}_j(t),\bm{\sigma}^{i}_j(t)), \ j = 1,2,\cdots,L,
\end{aligned}
\end{equation}
where $\bm{h}^{i}_j(t)$ represents the $j$-th latent factor learned by the input time series with a sampling rate of $r_i$ (i.e., the $j$-th latent factor in the $i$-th signal component) at time step $t$.
We use a dilation convolution operation to directly implement the function composition $\phi^{(i)} \circ s$ with appropriate padding to hold the same size between $\bm{X}^{(i)}$ and $\bm{X}$. 
The $\bm{\mu}^{i}_j(t)$ and $\bm{\sigma}^{i}_j(t)$ represent the $j$-th element of $\bm{\mu}^{i} \in \mathbb{R}^{L\times T}$ and $\bm{\sigma}^{i} \in \mathbb{R}^{L\times T}$ at time step $t$, respectively.
So 
$\mathcal{N}(\bm{\mu}^{i}_j(t),\bm{\sigma}^{i}_j(t))$ in Eq.(\ref{eq:encoder}) shows that the learned posterior distribution of the latent factor is a $1$-d Gaussian distribution with a mean value of $\bm{\mu}^{i}_j(t)$ and standard deviation of $\bm{\sigma}^{i}_j(t)$.

\subsection{Co-Attention-Weighted Decoder}

After encoding the historical observations into $K$ signal components, we present a co-attention-weighted decoder that includes two sets of sub-decoders with the same weight coefficients for interpolating the sub-decoders of each set.
The two sets aim to reconstruct the observation $\bm{X}$ and the prediction $\bm{Y}$, respectively.
The weights are set to be shared because both of them represent the importance of the signal components contributing to the formation of the observed data.

First, a linear transformation $f^{(i)}_{d}$ is used to reconstruct $\bm{\hat{X}}^{(i)}$ for each time scale separately, i.e., $\bm{\hat{X}}^{(i)} = f^{(i)}_{d}(\bm{h}^{i})$.
We calculate the weights of the sub-decoders according to the reconstruction of the original historical data rather than the predictions to avoid unnecessary error accumulation.
An attention mechanism is used to combine the reconstructed signals at different time scales, which is formulated as:
\begin{equation}
\begin{aligned}
\label{eq:decoder}
  & \alpha_{i} = \frac{\exp(-\|\bm{X}-\bm{\hat{X}}^{(i)}\|_{F}^{1/2})}{\sum_{i=1}^{K}\exp(-{\|\bm{X}-\bm{\hat{X}}^{(i)}\|_{F}^{1/2}})}, \\
  & \bm{\hat{X}} = \sum_{i=1}^{K}\alpha_i \bm{\hat{X}}^{(i)},
\end{aligned}
\end{equation}
where $\alpha_i$ is the attention-based weight coefficient, and $\|\cdot\|_{F}$ is the Frobenious norm of a matrix.
Gathering the linear transformation in each signal component, the decoder set for reconstructing the observation is described as $\psi_d=\{f^{(i)}_d\}|_{i=1}^{K}$, and we further have $\bm{\hat{X}} = \psi_d(\bm{H})$.

Second, for each signal component, an RNN-based sequence model $g^{(i)}$ is introduced to derive the next hidden state $\bm{\hat{h}}^{i}(t+1) = g^{(i)}(\bm{h}^{i}(j) |_{j=t-\epsilon+1}^{t})$ given the historical time series before time step $t+1$, where $\epsilon \in \mathbb{N}^{+}$ denotes the length of time steps needed to predict the former latent factors.
By using $\bm{\hat{h}}^{i}(t+1)$ and the shared weight coefficient $\alpha_i$, the final task-specified prediction can be calculated as
\begin{equation}
\begin{aligned}
\label{eq:predictor}
  \bm{\hat{Y}}(t+1) = \sum_{i=1}^{K}\alpha_i f^{(i)}_{y}(\bm{\hat{h}}^{i}(t+1)),
\end{aligned}
\end{equation}
where $f^{(i)}_{y}$ is designed as an arbitrary neural network for the downstream prediction task.
The decoder set for reconstructing the final prediction is thus formulated as $\psi_f = \{f^{(i)}_y\}|_{i=1}^{K}$, and we further have $\bm{\hat{Y}} = \psi_f(\bm{\hat{H}})$.

\section{Model Learning}

\label{sec:PLFl}

The proposed model encodes $\bm{X}$ into probability distributions of predictable latent factors $\bm{H}$, and $\bm{H}$ is used to generate the predictions of downstream tasks.
The encoded factors are encouraged to have the following three characteristics:

\begin{itemize}[leftmargin=1.25em]
    \item{Predictability.} As our presumption, there exist predictable latent factors underneath the unpredictable non-stationary time series. 
    \item{Sufficiency.} The sufficiency means the encoded factors contain enough information for the reconstruction and prediction.
    \item{Identifiability.} An identifiable model guarantees the correctness of drawing the joint distribution of the observations and latent signals, where the latent factors determine the trend of the observed time series.
\end{itemize}

\subsection{Learning Objective}
To achieve this goal, we propose a learning strategy that jointly maximizes the marginal probability of the proposed graphical model via variational inference and minimizes the task-specific objective.
Beforehand, an extra observed variable $\bm{E}$ is introduced to hold the identifiability of the whole model and disentangle the learned latent factors.
We define $\bm{E} = [\bm{e}_{1},\cdots,\bm{e}_{T}] \in \mathbb{R}^{(D+1)\times T}$ with $\bm{e}_{t}$ describing the joint representation of the time step $t$ and the corresponding observed signal $\bm{x}_{t}$.
With the help of $\bm{E}$, the prior distribution $p(\bm{H}|\bm{E})$ is assumed to be conditionally factorial, where all the latent factors $\bm{h}_j$ are independent each other.
In our model, all of the $\{\phi^{(i)}\}|_{i=1}^{K}$ jointly model the learned posterior distribution $q_\phi(\bm{H}|\bm{X},\bm{E})$ with $\phi$ as a parameter, which is used as a variation approximation to the true posterior distribution $p_\psi(\bm{H}|\bm{X},\bm{E})$ calculated by taking a marginal distribution of the joint probability $p_\psi(\bm{H},\bm{X}|\bm{E})$.
$\psi$ represents the total parameters of the decoder $\psi_d$ and the parameters of prior distributions $\psi_p$, so the joint distribution is described as:
\begin{equation}
\begin{aligned}
\label{eq:joint_distr}
  p_\psi(\bm{H},\bm{X}|\bm{E}) = p_{\psi_d}(\bm{X}|\bm{H}) p_{\psi_p}(\bm{H}|\bm{E}).
\end{aligned}
\end{equation}

Our model optimizes the parameters by maximizing the log-likelihood function:
\begin{equation}
\begin{aligned}
\label{eq:likelihood_abs}
  \max  &\sum_{(\bm{X},\bm{Y},\bm{E})\in \mathcal{D}}  \log p(\bm{X}|\bm{E})+\lambda \log p(\bm{Y}|\bm{X}),
\end{aligned}
\end{equation}
where $\mathcal{D}$ is the training set, and $\bm{Y}$ varies with the prediction task, \emph{i.e.}, the future observed data in the long-term series forecasting task, and the price change rate in the stock trend forecasting task.
The two terms in Eq.(\ref{eq:likelihood_abs}) focus on different aspects: $\log p(\bm{X}|\bm{E})$ aims to reconstruct $\bm{X}$ so that an informative representation $\bm{H}$ can be learned, and $\log p(\bm{Y}|\bm{X})$ serves for predications to encourage $\bm{H}$ to be predictable and discriminative, and $\lambda$ is a hyper-parameter to tune the relative importance of the two terms.
In this way, our method gains the advantages of both the generative and discriminative models to learn good representations for the specific forecasting task.

Eq.(\ref{eq:likelihood_abs}) is equivalent to
\begin{equation}
\begin{aligned}
\label{eq:likelihood-x}
  \max_{\phi,\psi,\upsilon,\psi_f} & \sum_{(\bm{X},\bm{Y},\bm{E})\in \mathcal{D}}  \mathbb{E}_{q_{\phi}(\bm{H}|\bm{X},\bm{E})} \Big[ \lambda \log  p_{\upsilon, \psi_f}(\bm{Y}|\bm{H}) + \\
  & \log p_{\psi}(\bm{H},\bm{X}|\bm{E})- \log q_{\phi}(\bm{H}|\bm{X},\bm{E}) \Big],
\end{aligned}
\end{equation}
where $\upsilon = \{g^{(i)}\}|_{i=1}^K$ denotes the total parameters of the predictive sequence models for hidden states, and $\upsilon$ together with $\psi_f$ form the task-specific prediction networks. 
The deduction from Eq.(\ref{eq:likelihood_abs}) to Eq.(\ref{eq:likelihood-x}) is elaborated in the Appendix.

Under the assumption of conditionally factorial, we show that Eq.(\ref{eq:likelihood-x}) can further be expanded as
\begin{equation}
\begin{aligned}
\label{eq:likelihood}
  \max_{\phi,\psi,\upsilon,\psi_f}   &\sum_{(\bm{X},\bm{E})\in \mathcal{D}} \big[\lambda \mathbb{E}_{q_{\phi}(\bm{H}|\bm{X},\bm{E})}\log  p_{\upsilon,\psi_f}(\bm{Y}|\bm{H})\\
  &+\sum_{i=1}^{K} \mathbb{E}_{q_{\phi^{(i)}}(\bm{h}^{i}|\bm{X}^{(i)},\bm{E})} \big( \log   p_{\psi^{(i)}}(\bm{h}^{i},\bm{X}^{(i)}|\bm{E})\\
  &- \log q_{\phi^{(i)}}(\bm{h}^{i}|\bm{X}^{(i)},\bm{E}) \big)\big].
\end{aligned}
\end{equation}
We provide details of the transformation procedure of Eq.~(\ref{eq:likelihood-x}) and Eq.~(\ref{eq:likelihood}) in the Appendix.
In the following, we give further explanation of Eq.~(\ref{eq:likelihood}) with respect to the three characteristics.

\subsection{Predictability}
\label{sec:predic}

Given a 1-D sequence $\bm{s} = \{\bm{s}(1),\bm{s}(2),\cdots,\bm{s}(t),\cdots\}$ where $t \in \mathbb{N}^{+}$ is the time step and $\bm{s}(t) \in \mathbb{R}$, the predictability of $\bm{s}$ is intuitively defined as

\textit{Definition 1} For any $\tau>0$, there exist $\epsilon, \delta \in \mathbb{N}^{+}$ and a function $g: \mathbb{R}^{\epsilon} \rightarrow \mathbb{R}$ which satisfies that the set $\{\bm{s}\in \mathbb{R}^{ \epsilon}|\lim_{\bm{s}\rightarrow \bm{s}^{'}} g(\bm{s})\neq g(\bm{s}^{'}) \}$ has measure zero, and for any $t > \epsilon$, 
\begin{equation}
\begin{aligned}
\label{eq:predicability}
  |g(\bm{s}(i)|_{i=t-\epsilon+1}^{t}) - \bm{s}(t+\delta)|<\tau,
\end{aligned}
\end{equation}
where $\tau$ is a small error tolerance when the sequence $\bm{s}$ is predictable.

{Eq. (\ref{eq:predicability}) suggests that historical data of a time series can be transformed into future data through continuous mapping, which is independent of time steps. 
If such a mapping exists, it indicates predictability within the time series data. 
This mapping is continuous to ensure that predicted values remain stable and do not undergo significant changes due to minor fluctuations in historical data. 
The mapping is independent of time steps, implying that the time series data itself exhibits specific patterns.}
According to the definition, the function $g$ in Eq.~(\ref{eq:predicability}) is essential to the predictability of $\bm{x}$, but it is often difficult to find or design.
As described in Eq.~(\ref{eq:predictor}), we propose to use a sequential neural network to approximate $g$ for endowing the latent factors with predictability. 
For each signal component, an RNN-based model $g^{(i)} (i=1,2,\cdots,K)$ is introduced and $|\bm{\hat{h}}^{i}(t+1) - \bm{h}^{i}(t+1)|<\tau$ is expected where $\bm{\hat{h}}^{i}(t+1)$ is the output of the model.
In general supervised learning, $\bm{h}^{i}(t+1)$ is treated as the ground-truth for training $g^{(i)}$.
However, $\bm{h}^{i}(t+1)$ is calculated by the encoder $\phi^{(i)}$, and it is undetermined before the end-to-end training, which might degenerates the leaning of $g$.
Instead, both of $\bm{h}^{i}(t+1)$ learned by $\phi^{(i)}$ and $\bm{\hat{h}}^{i}(t+1)$ predicted by $g^{(i)}$ are used to reconstruct the original historical and future signals $\bm{X}$, respectively, that are determinate and common.
We show that feasibility is warranted in the following result. 

\textit{Theorem 1} For any $\tau_1, \tau_2>0$, there exist $\epsilon, \delta \in \mathbb{N}^{+}$, a function $g: \mathbb{R}^{\epsilon} \rightarrow \mathbb{R}$ which satisfies that the set $\{\bm{s}\in \mathbb{R}^{\epsilon}|\lim_{\bm{s}\rightarrow \bm{s}^{'}} g(\bm{s})\neq g(\bm{s}^{'}) \}$ has measure zero, and a continuous and derivable function $l: \mathbb {R} \rightarrow \mathbb{R}$, for any $t > \epsilon$,
\begin{equation}
\begin{aligned}
\label{eq:predicability_theo}
  |l(g(\bm{s}(i)|_{i=t-\epsilon+1}^{t}))| < \tau_1,  \ |l(\bm{s}(t+\delta))|<\tau_2
\end{aligned}
\end{equation}
holds, then $\bm{s}$ must be predictable.

\begin{proof}
We rewrite $g(\bm{s}(i)|_{i=t-\epsilon+1}^{t})$ as $\hat{s}$ for simplicity.
Without loss of generality, $\hat{s}$ is assumed to be less than $\bm{s}(t+\delta)$.
Since that $l$ is continuous and derivable, there must exist $\hat{s}<\xi<\bm{s}(t+\delta)$ such that
\begin{equation}
\begin{aligned}
\label{eq:lmvt}
  l^{'}(\xi)(\bm{s}(t+\delta)-\hat{s}) = l(\bm{s}(t+\delta))-l(\hat{s}),
\end{aligned}
\end{equation}
according to the Lagrange mean value theorem.
For any $\tau > 0$, let $\tau_1=\tau_2= \tau \times |l{'}(\xi)|/2$, then we have $|l(\hat{s})| < \tau_1$ and $|l(\bm{s}(t+\delta))|<\tau_2$.
Accordingly,
\begin{equation}
\begin{aligned}
\label{eq:uneq_02}
  |l(\hat{s}) - l(\bm{s}(t+\delta))| &\leq |l(\hat{s})| + |l(\bm{s}(t+\delta))| \\
  &< \tau_1+\tau_2 = \tau\times |l{'}(\xi)|.
\end{aligned}
\end{equation}
Substituting Eq.(\ref{eq:lmvt}) into (\ref{eq:uneq_02}), we thus have that
\begin{equation}
\begin{aligned}
\label{eq:uneq_03}
  |l^{'}(\xi)||(\bm{s}(t+\delta)-\hat{s})| < \tau\times|l{'}(\xi)|,
\end{aligned}
\end{equation}
and further $|g(\bm{s}(i)|_{i=t-\epsilon+1}^{t}) - \bm{s}(t+\delta)|<\tau$.
Therefore, $\bm{s}$ is predictable.
\end{proof}

\subsection{Sufficiency}
\label{sec:suff}

The learned components should be informative enough for both the forecasting task and reconstructing $\bm{X}$, and it is implemented by learning the proposed co-attention-weighted decoder that should be designed according to specific time series forecasting tasks.
In this paper, we present two typical tasks as examples to show the prediction learning ability of our model, i.e., long-term series forecasting and stock trend forecasting.

\noindent{\textbf{Long-term series forecasting.}}
The goal of this task is to predict the future $H$-horizon data $\bm{Y}=[\bm{x}(T+1),\bm{x}(T+2),\dots,\bm{x}(T+H)] \in \mathbb{R}^{D \times H}$ given historical data before time step $T+1$.
We apply the non-dynamic decoder to our method for generating $\bm{Y}$.
To be specific, an MLP is used as $f_y^{(i)}$ to calculate all the $H$-horizon prediction of the $i$-th sampled time series given the predicted hidden state $\bm{\hat{h}}^{i}(T+1)$.
We use the mean squared error (MSE) loss to guide the learning of this characteristic with respect to the reconstruction of future data (i.e., forecasting). 
In this task, the output prediction is just the future data of the original time series, which is known as the auto-regressive generation task, so this objective is equivalent to learning the characteristic of predictability.

\noindent{\textbf{Stock trend forecasting.}}
Different from the aforementioned auto-regressive generation task, stock trend forecasting aims to predict the stock price change rate given some stock-related variables (e.g., the stock price at specific times and the trading volume).
The prediction $\bm{Y}$ is defined as $\bm{Y}(T+1) = (P(T+1)-P(T))/P(T)$ where $P$ is the real price.
It is obvious that $\bm{Y}$ is defined in two consecutive time steps, but sampling for multiple signal components would break the consecutiveness of $\bm{h}^{i}$.
Fortunately, the designed function $g$ ensures $\bm{h}^{i}$ holding the predictability, so the predicted $\hat{\bm{h}}^{(i)}$ is just capable of filling the time gaps.
Suppose that the latent components $\bm{h}^{i}$ learned by using the sampled $\bm{X}^{(i)}$ are represented as $\{\cdots,\bm{h}^{i}(t_{1}), \bm{h}^{i}(t_{2}), \dots,\bm{h}^{i}(T)\}$ where $t_{k}-t_{k-1} = \eta$ is a constant, then we use $g^{(i)}$ to predict $\{\cdots,\hat{\bm{h}}^{(i)}(t_{1}+1), \hat{\bm{h}}^{(i)}(t_{2}+1), \dots,\hat{\bm{h}}^{(i)}(T+1)\}$, and combine them together to learn representations for the forecasting task, i.e., 
\begin{equation}
\begin{aligned}
\label{eq:combine}
  \{&\cdots,(\bm{h}^{i}(t_{1}),\hat{\bm{h}}^{(i)}(t_{1}+1)), (\bm{h}^{i}(t_{2}),\hat{\bm{h}}^{(i)}(t_{2}+1)), \dots,\\
  &(\bm{h}^{i}(T), \hat{\bm{h}}^{(i)}(T+1)) \}.
\end{aligned}
\end{equation}
Using representations of Eq.(\ref{eq:combine})  as inputs, another RNN-based model is used as $f_y^{(i)}$ to predict $\bm{Y}$.
As shown in Eq.(\ref{eq:predictor}), the $\alpha_{i}$ in Eq.(\ref{eq:decoder}) is used to integrate the outputs of sequence models in various signal components.
An MSE regression loss between the predicted price return and the ground-truth is calculated as the guidance of factor learning for keeping the sufficiency of information.

Furthermore, for both tasks, the sub-decoders composed of linear functions for reconstruction are designed as simple as possible to enforce $\bm{H}$ informative enough for reconstructing $\bm{X}$.
However, the predictability and identifiability require an injective decoder which is often approximated by a complex neural network.
The contradiction is alleviated by introducing the sum-injective theorem \cite{DBLP:conf/iclr/XuHLJ19} as shown in Sec.~\ref{sec:identifi}.

\subsection{Identifiability}
\label{sec:identifi}

The identifiability of our model is ensured by Theorem 1 of \cite{DBLP:conf/aistats/KhemakhemKMH20}.
In order to fit the theorem, the prior distribution should be elaborated.
Different from the standard normal distribution prior used in the classical VAE, the Gaussian location-scale family is chosen as the prior distribution $p_{\psi_p}(\bm{h^{(i)}_j}|\bm{E})$, which is rewritten in the form of exponential family distribution as 
\begin{equation}
\begin{aligned}
\label{eq:exp_prior}
  &\frac{1}{\sqrt{2\pi}\sigma} \exp\big(-\frac{\mu^2}{2\sigma^2}\big) \exp\big( \begin{bmatrix} \bm{h}^{i}_j & (\bm{h}^{i}_j)^{2} \end{bmatrix}^{\top} \begin{bmatrix} \frac{\mu}{\sigma^2} \\ -\frac{1}{2\sigma^2} \end{bmatrix}\big), \\
\end{aligned}
\end{equation}
where $\mathcal{T}^{(i)}(\bm{h}^{i}_j) = (\bm{h}^{i}_j, (\bm{h}^{i}_j)^{2})$ are sufficient statistics, and the corresponding parameters $\lambda^{(i)}(\bm{E}) = (\frac{\mu}{\sigma^2}, -\frac{1}{2\sigma^2})$ and the normalizing constant $\frac{1}{\sqrt{2\pi}\sigma} \exp\big(-\frac{\mu^2}{2\sigma^2}\big)$ are dependent on $\bm{E}$.
In our model, the $\lambda^{i}$ are independently and randomly sampled according to $\bm{E}$ to achieve the aforementioned characteristic of conditionally independent of the latent factors, and $\mathcal{T}^{(i)}$ are same for different components. 
Keeping consistent with \cite{DBLP:conf/aistats/KhemakhemKMH20}, the parameters of the joint distribution are $ \psi = \{\psi_d,\psi_p\}$ where $\psi_d=\{f_c^{(i)}\}$ and $\psi_{p} = \{\mathcal{T}^{(i)}, \lambda^{(i)}\}$ when gathering the parameters of all components.
For the deep latent variable model, the $\sim_A$ identifiable is introduced as

\textit{Definition 2} Define $\sim$ as the equivalence relation: 
\begin{equation}
\begin{aligned}
\label{eq:equivalence}
  (f, \mathcal{T}, \lambda) &\sim (\tilde{f},\widetilde{\mathcal{T}},\tilde{\lambda}) \Leftrightarrow \\
  \exists \bm{A}, c | \mathcal{T}(f^{-1}(\bm{x})) &= \bm{A}\widetilde{\mathcal{T}}(\tilde{f}^{-1}(\bm{x}))+\bm{c}, \forall x
\end{aligned}
\end{equation}
where $\bm{A}$ is an $LK \times LK$ matrix, $\bm{c}$ is a vector.
If $\bm{A}$ is invertible, this relation is denoted as $\sim_A$.

Using the theorem, our deep latent factor model is $\sim_A$ identifiable since four conditions are satisfied:
\begin{enumerate}[leftmargin=1.25em]
    \item The decoder $f_d$ is with zero measure error.
    \item The $f_d$ is composed of fully connected layers and non-linear attention units, and thus not injective. Fortunately, we can assert that our attention-based structure is learned to approximate an injective according to the researches \cite{DBLP:conf/iclr/XuHLJ19} and \cite{wijesinghe2021new} on the aggregation operation of graph attention networks. According to \cite{wijesinghe2021new}, the sum aggregation in Eq.(\ref{eq:decoder}) is adjusted by adding $1$ to $\alpha$ to ensure the injective.
    \item Obviously, $\mathcal{T}$ is differentiable everywhere, and $\bm{h}_{j}^{(i)}$ and $(\bm{h}^{i}_j)^{2}$ are linearly independent.
    \item There exist $LK+1$ points $\bm{E}_i|_{i=1}^{LK+1}$ such that the matrix $[\lambda(\bm{E}_2)-\lambda(\bm{E}_1),\cdots,\lambda(\bm{E}_{LK+1})-\lambda(\bm{E}_{LK})]$ is invertible, which is achieved by randomly and independently sampling $\mu$ and $\sigma$ according to $\bm{E}$.
\end{enumerate}
{Overall, with the introduction of temporal structure as advocated in \cite{DBLP:conf/aistats/KhemakhemKMH20}, we demonstrate that our model possesses identifiability. 
Similar to non-linear ICA models \cite{DBLP:conf/nips/HalvaCLSZGH21,DBLP:conf/icml/LocatelloBLRGSB19,DBLP:conf/nips/HyvarinenM16}, identifiability implies that the model can uniquely infer its parameters or representations from the observed data. 
In other words, the learned model contains sufficient information to differentiate different factors or features, thus having the capability to generate disentangled representations. 
Furthermore, the disentanglement of latent factors leads to sparse connections in our model, consequently reducing the model's complexity and facilitating more efficient learning.}

\begin{table*}[tbp]
  \centering
  \caption{Comparisons on the long-term series forecasting performances of the existing and the proposed methods. Values marked in bold and with underlines represent the best and the second-best scores on the corresponding datasets, respectively.}
    \resizebox{\textwidth}{!}{
    \begin{tabular}{ll|cccccccccccccccccc}
\toprule
\multirow{2}{*}{}                         &     & \multicolumn{2}{c}{Ours}  & 
\multicolumn{2}{c}{\makecell[c]{PatchTST\\ \cite{DBLP:conf/iclr/NieNSK23} }} & 
\multicolumn{2}{c}{\makecell[c]{DLinear\\ \cite{DBLP:conf/aaai/ZengCZ023} }} & 
\multicolumn{2}{c}{\makecell[c]{N-HITS\\ \cite{DBLP:journals/corr/abs-2201-12886} }}      &  
\multicolumn{2}{c}{\makecell[c]{NS Transformer\\ \cite{liunon} }}&
\multicolumn{2}{c}{\makecell[c]{FEDformer\\ \cite{DBLP:conf/icml/ZhouMWW0022} }} & 
\multicolumn{2}{c}{\makecell[c]{LSTNet\\ \cite{DBLP:conf/sigir/LaiCYL18} }}  & 
\multicolumn{2}{c}{\makecell[c]{DilRNN\\ \cite{DBLP:conf/nips/ChangZHYGTCWHH17} }}  & 
\multicolumn{2}{c}{\makecell[c]{ARIMA\\ \cite{anderson1976time} }} \\
                                          & H   & MSE                                & MAE                                & MSE            & MAE            & MSE              & MAE             & MSE        & MAE              & MSE            & MAE           & MSE            & MAE            & MSE          & MAE         & MSE          & MAE         & MSE          & MAE        \\ \midrule
\multirow{4}{*}{\rotatebox{90}{ETTm$_2$}} & 96  & \textbf{0.135} & \textbf{0.203} & \underline{0.166} & 0.256 & 0.167 & 0.260 & 0.176 & \underline{0.255} & 0.192 & 0.274 & 0.203 & 0.287 & 3.142 & 1.365 & 0.343 & 0.401 & 0.225 & 0.301 \\
                                          & 192 & \textbf{0.213} & \textbf{0.271} & \underline{0.223} & \underline{0.296}  & 0.224 & 0.303 & 0.245 & 0.305 & 0.280 & 0.339 & 0.269 & 0.328 & 3.154 & 1.369 & 0.424 & 0.468 & 0.298 & 0.345 \\
                                          & 336 & \textbf{0.257} & \textbf{0.324} & \underline{0.274} & \underline{0.329} & 0.281 & 0.342 & 0.295 & 0.346 & 0.334 & 0.361 & 0.325 & 0.366 & 3.160 & 1.369 & 0.632 & 1.083 & 0.370 & 0.386 \\
                                          & 720 & \textbf{0.357} & \textbf{0.385} & \underline{0.361} & \textbf{0.385} & 0.397 & 0.421 & 0.401 & \underline{0.413} & 0.417 & \underline{0.413} & 0.421 & 0.415 & 3.171 & 1.368 & 0.634 & 0.594 & 0.478 & 0.445 \\ \midrule
\multirow{4}{*}{\rotatebox{90}{ECL}}      & 96  & \textbf{0.127} & \textbf{0.219} & \underline{0.129} & \underline{0.222} & 0.140 & 0.237 & 0.147 & 0.249 & 0.169 & 0.273 & 0.183 & 0.297 & 0.680 & 0.645 & 0.233 & 0.927 & 1.220 & 0.814 \\
                                          & 192 & \textbf{0.145} & \underline{0.241} & \textbf{0.147} & \underline{0.240} & 0.153 & 0.249 & 0.167 & 0.269 & 0.182 & 0.286 & 0.195 & 0.308 & 0.725 & 0.676 & 0.265 & 0.921 & 1.264 & 0.842 \\
                                          & 336 & \textbf{0.160} & \textbf{0.255} & \underline{0.163} & \underline{0.259} & 0.169 & 0.267 & 0.186 & 0.290 & 0.200 & 0.304 & 0.212 & 0.313 & 0.828 & 0.727 & 0.235 & 0.896 & 1.311 & 0.866 \\
                                          & 720 & \textbf{0.197} & \textbf{0.287} & \textbf{0.197} & \underline{0.290} & \underline{0.203} & 0.301 & 0.243 & 0.340 & 0.222 & 0.321 & 0.231 & 0.343 & 0.957 & 0.811 & 0.322 & 0.890 & 1.364 & 0.891 \\ \midrule
\multirow{4}{*}{\rotatebox{90}{Exchange}} & 96  & \multicolumn{1}{l}{\textbf{0.065}} & \textbf{0.162} & - & - & \underline{0.081} & 0.203 & 0.092 & \underline{0.202} & 0.111 & 0.237 & 0.139 & 0.276 & 1.551 & 1.058 & 0.383 & 0.450 & 0.296 & 0.214 \\
                                          & 192 & \textbf{0.126} & \textbf{0.257} & - & - & \underline{0.157} & \underline{0.293} & 0.208 & 0.322 & 0.219 & 0.335 & 0.256 & 0.369 & 1.477 & 1.028 & 1.123 & 0.834 & 1.056 & 0.326 \\
                                          & 336 & \textbf{0.181} & \textbf{0.335} & - & - & 0.305 & 0.414 & \underline{0.301} & \underline{0.403} & 0.421 & 0.476 & 0.426 & 0.464 & 1.507 & 1.031 & 1.612 & 1.051 & 2.298 & 0.467 \\
                                          & 720 & \textbf{0.476} & \textbf{0.515} & - & - & \underline{0.643} & 0.601 & 0.798 & \underline{0.596} & 1.092 & 0.769 & 1.090 & 0.800 & 2.285 & 1.243 & 1.827 & 1.131 & 20.666 & 0.864 \\ \midrule
\multirow{4}{*}{\rotatebox{90}{Traffic}} & 96  & \textbf{0.356} & \textbf{0.239} & \underline{0.360} & \underline{0.249} & 0.410 & 0.282 & 0.402 & 0.282 & 0.612 & 0.338 & 0.562 & 0.349 & 1.107 & 0.685 & 0.580 & 0.308 & 1.997 & 0.924 \\
                                          & 192 & \textbf{0.375} & \textbf{0.250} & \underline{0.379} & \underline{0.256} & 0.423 & 0.287 & {0.420} & {0.297} & 0.613 & 0.340 & 0.562 & 0.346 & 1.157 & 0.706 & 0.739 & 0.383 & 2.044 & 0.944 \\
                                          & 336 & \textbf{0.390} & \underline{0.266} & \underline{0.392} & \textbf{0.264} & 0.436 & 0.296 & 0.448 & 0.313 & 0.618 & 0.328 & 0.570 & 0.323 & 1.216 & 0.730 & 0.804 & 0.419 & 2.096 & 0.960 \\
                                          & 720 & \textbf{0.427} & \textbf{0.283} & \underline{0.432} & \underline{0.286} & 0.466 & 0.315 & {0.539} & {0.353} & 0.653 & {0.355} & 0.596 & 0.368 & 1.481 & 0.805 & 0.695 & 0.372 & 2.138 & 0.971 \\ \midrule
\multirow{4}{*}{\rotatebox{90}{Weather}}  & 96  & \textbf{0.145} & \textbf{0.187} & \underline{0.149} & 0.198 & 0.176 & 0.237 & 0.158 & \underline{0.195} & 0.173 & 0.223 & 0.217 & 0.296 & 0.594 & 0.587 & 0.193 & 0.245 & 0.217 & 0.258 \\
                                          & 192 & \textbf{0.194} & \textbf{0.240} & \textbf{0.194} & \underline{0.241} & 0.220 & 0.282 & \underline{0.211} & 0.247 & 0.245 & 0.285 & 0.276 & 0.336 & 0.560 & 0.565 & 0.255 & 0.306 & 0.263 & 0.299 \\
                                          & 336 & \underline{0.246} & \textbf{0.278} & \textbf{0.245} & \underline{0.282} & 0.265 & 0.319 & 0.274 & {0.300} & 0.321 & 0.338 & 0.339 & 0.380 & 0.597 & 0.587 & 0.329 & 0.360 & 0.330 & 0.347 \\
                                          & 720 & \textbf{0.313} & \textbf{0.330} & \underline{0.314} & \underline{0.334} & 0.323 & 0.362 & {0.351} & {0.353} & 0.414 & 0.410 & 0.403 & 0.428 & 0.618 & 0.599 & 0.521 & 0.495 & 0.425 & 0.405 \\ \midrule
\multirow{4}{*}{\rotatebox{90}{ILI}}      & 24  & \multicolumn{1}{l}{\textbf{0.946}} & \multicolumn{1}{l}{\textbf{0.559}} & \underline{1.319} & \underline{0.754} & 2.215 & 1.081 & 1.862 & {0.869} & 2.294 & 0.945 & 2.203 & 0.963 & 6.026 & 1.770 & 4.538 & 1.449 & 5.554 & 1.434 \\
                                          & 36 & \textbf{0.997} & \textbf{0.636} & \underline{1.579} & 0.870 & 1.963 & 0.963 & 2.071 & 0.934 & 1.825 & \underline{0.848} & 2.272 & 0.976 & 5.340 & 1.668 & 3.709 & 1.273 & 6.940 & 1.676 \\
                                          & 48  & \textbf{1.107} & \textbf{0.734} & \underline{1.553} & \underline{0.815} & 2.130 & 1.024 & 2.134 & 0.932 & 2.010 & {0.900} & 2.209 & 0.981 & 6.080 & 1.787 & 3.436 & 1.238 & 7.192 & 1.736 \\
                                          & 60  & \multicolumn{1}{l}{\textbf{1.260}} & \textbf{0.753} & \underline{1.470} & \underline{0.788} & 2.368 & 1.096 & 2.137 & 0.968 & 2.178 & {0.963} & 2.545 & 1.061 & 5.548 & 1.720 & 3.703 & 1.272 & 6.648 & 1.656 \\ \bottomrule
\end{tabular}}
  \label{tab:main_result2}%
\end{table*}%

\section{Experiments}

\subsection{Settings}
\subsubsection{Datasets}

The proposed method\footnote{https://github.com/houjingyi-ustb/discover\_PLF} is evaluated on two typical time series forecasting tasks, i.e., long-term series forecasting and stock trend forecasting.

For long-term series forecasting, we conduct experiments on 6 public datasets:
(1) The ETTm$_2$ \cite{DBLP:conf/aaai/ZhouZPZLXZ21} dataset records a 7-dimensional feature including oil temperature and loads of an electricity transformer every 15 minutes.
(2) The Exchange \cite{DBLP:conf/sigir/LaiCYL18} dataset includes the daily exchange rates of 8 countries.
(3) The \href{https://archive.ics.uci.edu/ml/datasets/ElectricityLoadDiagrams20112014}{ECL} dataset collects hourly electricity consumption from 321 customers.
(4) The \href{http://pems.dot.ca.gov}{Traffic} dataset includes hourly road occupancy rates collected from 862 sensors.
(5) The \href{https://www.bgc-jena.mpg.de/wetter}{Weather} dataset reports 21 meteorological indicators (i.e., air temperature, humidity, etc.) every 10 minutes from a weather station.
(6) The \href{https://gis.cdc.gov/grasp/fluview/fluportaldashboard.html}{ILI} records weekly ratios of influenza-like illness patients.
We follow standard protocols, chronologically splitting ETTm$_2$ into training, validation, and test sets by the ratio of $7:1:2$ and other datasets by the ratio of $6:2:2$.
The horizon lengths are set to $H \in \{96, 192, 336, 720\}$ for the first 5 datasets, and $H \in \{24,36,48,60\}$ for ILI.
For the evaluation, we use the mean absolute error (MAE) and the mean squared error (MSE) metrics.

For stock trend forecasting, we conduct experiments on CSI100 and CSI300 in the publicly available stock dataset, Alpha360 \cite{DBLP:journals/corr/abs-2009-11189}.
The input data of this task are the sequences of 6-dimensional stock prices in 60 trading days and the output data are the price returns of the next day.
For fair comparisons, we follow the experimental settings of \cite{DBLP:journals/corr/abs-2110-13716}.  
We split the data from both sets according to temporal order, where the training, validation, and test data are sampled in the range of 01/01/2007 to 12/31/2014, 01/01/2015 to 12/31/2016, and 01/01/2017 to 12/31/2020, respectively.
As for the evaluation metrics, the popular Information Coefficient (IC), Rank IC, and Precision@$N$ (P@$N$) with $N \in \{3,5,10,30\}$ for quantitative investment are employed.
We calculate the mean and standard deviation of the results by repeating each experiment 10 times.

\subsubsection{Implementation Details}

For encoders with different sampling rates, we use 1D convolutional operations with different dilation rates to capture local contexts in temporal sequences. 
After comparing the validation results, we get the optimal setting, that is, factorizing the original time series into 2 or 3 signal components with different sampling rates corresponding to specific datasets, further analyzed in section \ref{sub:ablation}.
The convolution kernel size is 3.
We pad $(k-1)\times r$-dimensional zero vectors before the very first of the input features to maintain length.
Residual connections via linear projections are applied to the convolution layers for better optimization.
The hidden size of each latent component is set to 128.
The number of the hidden units is empirically set to be 128 according to the prior related work.
The coefficients of losses are selected from $\{1,1e1,1,5e-1,1e-1\}$ according to the corresponding performances, and the final coefficient for KL loss is determined to be $5e-1$, while others are set to $1$ for stock trend forecasting.

For long-term series forecasting, we follow the auto-tuning strategy of \cite{DBLP:journals/corr/abs-2201-12886}.
The learning rate is sampled between $[1e-4, 5e-4]$, and the input size is sampled from $\{k\times H|1\leq k \leq 10\}$.
We emphasize the prediction loss for long-term series forecasting to enable the model to concentrate more on generating relatively hard-to-predict long-term data by multiplying the losses with $1e-3$ except for the prediction loss.
Actually, from our experiments, the longer $H$ the better, but for computational sufficiency, the experimental results reported in this paper are done by limiting $H<2000$.
The sampling rates (dilation rates) of different datasets contain 1, and others are chosen from basic time units, such as hour, day, week, and year.
The sequence model is selected from $\{$LSTMs, GRUs$\}$ for different tasks. 
The task-specific decoder is a 2-layer MLP, where the top layer is for the prediction generation and the bottom layer has 128-dimensional nodes with the ReLU activation.
Because different variables at the same time step often represent homogeneous semantics in long-term forecasting datasets, we directly conduct univariate prediction following \cite{DBLP:journals/corr/abs-2201-12886,DBLP:conf/iclr/NieNSK23,DBLP:conf/aaai/ZengCZ023}.
{
Actually, our method is more suitable for multivariate inputs, so we embed each 1-dimensional time series into $\{8,16,32\}$-dimensional sequential data using the patching operation following \cite{DBLP:conf/iclr/NieNSK23} but setting the stride to 1.
}
To enable stable optimization, we perform data normalization within each mini-batch.
Specifically, before feeding data into the proposed network, we calculate the mean and variance of all data in the mini-batch and normalize using these statistics.
After generating predictions via the model, we reverse the normalization of predicted data by multiplying by the variance and adding the mean calculated initially to obtain the final prediction.

For stock trend forecasting, the learning rate is $2e-4$.
We simply apply a linear regression model to our method for end-to-end training (Ours+LR). 
We also cascade the knowledge-driven model, HIST \cite{DBLP:journals/corr/abs-2110-13716}, into our model to further improve the performance with the guidance of extra information from the market (Ours+HIST).
In the experiments of this task, we further improve the model performance by applying the disentanglement trick.
Eq.~(\ref{eq:likelihood}) can further be decomposed into a reconstruction term and a KL divergence between prior and posterior distributions, and the importance of the KL divergence can be gradually improved to disentangle the latent factors by tuning hyper-parameters during the training procedure of our model.
Here, the loss of the KL divergence is multiplied by a coefficient $\beta$ in addition to $5e-1$ according to the $\beta$-VAE \cite{DBLP:conf/iclr/HigginsMPBGBML17} for disentangling the predictive latent factors.
We gradually increase the value of $\beta$ in stages, which is
\begin{equation}
\label{eq6}
\beta=\left\{
\begin{aligned}
&0.1, \ \ \ \mathrm{epoch}<20,\\
&0.5, \ \ \ 20\leq \mathrm{epoch}<30,\\
&1, \ \ \ \mathrm{epoch}\geq 30.\\
\end{aligned}
\right.
\end{equation}
We observe that the optimization converges within 50 epochs.

\begin{table*}[tbp]
  \scriptsize
  \centering
  \caption{Test results (and the corresponding standard deviations) of the existing and the proposed methods on CSI100 and CSI300. The ``*'' indicates using extra knowledge. Values marked in bold and with underlines represent the highest scores with and without extra information, respectively.}
    \setlength{\tabcolsep}{3mm}{
    \begin{tabular}{l|cccccc|cccccc}
    \toprule
    \multirow{2}{*}{Methods} & \multicolumn{6}{c|}{CSI100}     & \multicolumn{6}{c}{CSI300} \\
\cmidrule{2-13}          & IC    & Rank IC  & P@3 & P@5 & P@10    & P@30   & IC    & Rank IC  & P@3 & P@5 & P@10    & P@30   \\
    \midrule
    \multirow{2}{*}{MLP}&0.071&0.067&56.53&56.17&55.49&53.55&0.082&0.079&57.21&57.10&56.75&55.56\\
    &(4.8e-3)&(5.2e-3)&(0.91)&(0.48)&(0.30)&(0.36)&(6e-4)&(3e-4)&(0.39)&(0.33)&(0.34)&(0.14)\\
    \midrule
    \multirow{2}{*}{LSTM \cite{DBLP:journals/neco/HochreiterS97}}&0.097&0.091&60.12&59.49&59.04&54.77&0.104&0.098&59.51&59.27&58.40&56.98\\
    &(2.2e-3)&(2.0e-3)&(0.52)&(0.19)&(0.15)&(0.11)&(1.5e-3)&(1.6e-3)&(0.46)&(0.34)&(0.30)&(0.11)\\
    \midrule
    \multirow{2}{*}{GRU \cite{DBLP:journals/corr/ChungGCB14}}&0.103&0.097&59.97&58.99&58.37&55.09&0.113&0.108&59.95&59.28&58.59&57.43\\
    &(1.7e-3)&(1.6e-3)&(0.63)&(0.42)&(0.29)&(0.15)&(1.0e-3)&(8e-4)&(0.62)&(0.35)&(0.40)&(0.28)\\
    \midrule
    \multirow{2}{*}{ALSTM \cite{DBLP:conf/uksim/AriyoAA14}}&0.102&0.097&\underline{60.79}&\underline{59.76}&58.13&55.00&0.115&0.109&59.51&59.33&58.92&57.47\\
    &(1.8e-3)&(1.9e-3)&(0.23)&(0.42)&(0.13)&(0.12)&(1.4e-3)&(1.4e-3)&(0.20)&(0.51)&(0.29)&(0.16)\\
    \midrule
    \multirow{2}{*}{SFM \cite{DBLP:conf/kdd/ZhangAQ17}}&0.081&0.074&57.79&56.96&55.92&53.88&0.102&0.096&59.84&58.28&57.89&56.82\\
    &(7.0e-3)&(8.0e-3)&(0.76)&(1.04)&(0.60)&(0.47)&(3.0e-3)&(2.7e-3)&(0.91)&(0.42)&(0.45)&(0.39)\\
    \midrule
    \multirow{2}{*}{GAT \cite{DBLP:conf/iclr/VelickovicCCRLB18}}&0.096&0.090&59.17&58.71&57.48&54.59&0.111&0.105&60.49&59.96&59.02&57.41\\
    &(4.5e-3)&(4.4e-3)&(0.68)&(0.52)&(0.30)&(0.24)&(1.9e-3)&(1.9e-3)&(0.39)&(0.23)&(0.14)&(0.30)\\
    \midrule
    \multirow{2}{*}{Transformer \cite{DBLP:conf/ijcai/DingWSGG20}}&0.089&0.090&59.62&59.20&57.94&54.80&0.106&0.104&60.76&60.06&59.48&57.71\\
    &(4.7e-3)&(5.1e-3)&(1.20)&(0.84)&(0.61)&(0.33)&(3.3e-3)&(2.5e-3)&(0.35)&(0.20)&(0.16)&(0.12)\\
    \midrule
    \multirow{2}{*}{TRA \cite{DBLP:conf/kdd/LinZL021}}&0.107&0.102&60.27&59.09&57.66&55.16&0.119&0.112&60.45&59.52&59.16&\underline{58.24}\\
    &(2.0e-3)&(1.8e-3)&(0.43)&(0.42)&(0.33)&(0.22)&(1.9e-3)&(1.7e-3)&(0.53)&(0.58)&(0.43)&(0.32)\\
    \midrule
    \multirow{2}{*}{FactorVAE \cite{duan2022factorvae}}&0.107&0.102&60.23&{59.51}&\underline{58.39}&55.21&0.118&0.110&61.04&60.73&59.26&57.90\\
    &(2.1e-3)&(1.5e-3)&(0.78)&(0.45)&(0.45)&(0.76)&(2.3e-3)&(3.1e-3)&(0.48)&(0.62)&(0.17)&(0.51)\\
    \midrule
    \multirow{2}{*}{HIST \cite{DBLP:journals/corr/abs-2110-13716}*}&0.120&0.115&61.87&60.82&59.38&56.04&0.131&0.126&61.60&61.08&60.51&58.79\\
    &(1.7e-3)&(1.6e-3)&(0.47)&(0.43)&(0.24)&(0.19)&(2.2e-3)&(2.2e-3)&(0.59)&(0.56)&(0.40)&(0.31)\\
    \midrule
    \multirow{2}{*}{\textbf{Ours+LR}}&\underline{0.111}&\underline{0.105}&{60.52}&59.60&58.09&\underline{55.31}&\underline{0.120}&\underline{0.113}&\underline{62.18}&\underline{61.09}&\underline{59.71}&{58.10}\\
    &(1.8e-3)&(1.5e-3)&(0.63)&(0.33)&(0.25)&(0.18)&(2.6e-3)&(2.4e-3)&(0.98)&(0.76)&(0.47)&(0.24)\\
    \midrule
    \multirow{2}{*}{\textbf{Ours+HIST}*}&\textbf{0.128}&\textbf{0.122}&\textbf{62.41}&\textbf{61.41}&\textbf{60.06}&\textbf{56.46}&\textbf{0.136}&\textbf{0.131}&\textbf{63.08}&\textbf{62.39}&\textbf{61.59}&\textbf{59.45}\\
    &(2.2e-3)&(1.9e-3)&(0.50)&(0.53)&(0.15)&(0.14)&(3.1e-3)&(3.2e-3)&(0.58)&(0.60)&(0.28)&(0.18)\\

    \bottomrule
    \end{tabular}}%
  \label{tab:main_result}%
\end{table*}%

\begin{table*}[tbp]
  \centering
  \caption{ Test results of ablation studies. The MP indicates the mean precision of Precision@$N$, where $N \in \{3,5,10,30\}$. The \checkmark and $\times$ represent conducting experiments with and without the corresponding module, respectively.}
    \resizebox{0.95\textwidth}{!}{
    \begin{tabular}{ccccc|ccc|ccc|cc|cc|cc}
    \toprule
    \multirow{2}{*}{Decomp.} &\multirow{2}{*}{Disent.} &\multirow{2}{*}{Reconst.}&\multirow{2}{*}{Ind.}&\multirow{2}{*}{S.W.} & \multicolumn{3}{c|}{CSI100}     & \multicolumn{3}{c|}{CSI300} & \multicolumn{2}{c|}{Exchange} & \multicolumn{2}{c|}{Weather}  & \multicolumn{2}{c}{ILI} \\
\cmidrule{6-17}   &  & &   &  & IC    & Rank IC  & MP  & IC    & Rank IC  & MP & MSE  & MAE  & MSE  & MAE  & MSE  & MAE  \\
    \midrule
    $\times$&$\times$&$\times$&$\times$&$\times$&0.103&0.098&58.15&0.113&0.108&58.92&0.289&0.397&0.251&0.310&2.089&0.977\\
    \midrule
    $\times$&\checkmark&\checkmark&$\times$&$\times$&0.105&0.100&58.26&0.116&0.110&58.96&0.285&0.366&0.246&0.303&1.661&0.808\\
    \midrule
    \checkmark&$\times$&\checkmark&\checkmark&\checkmark&0.108&0.103&58.23&0.119&0.111&60.23&0.213&0.318&0.232&0.271&1.236&0.692\\
    \midrule
    \checkmark&$\times$&$\times$&\checkmark&$\times$&0.106&0.102&58.42&0.117&0.111&60.03&0.213&0.319&0.242&0.274&1.456&0.697\\
    \midrule
    \checkmark&\checkmark&\checkmark&$\times$&$\times$&0.108&0.103&58.34&0.118&0.112&59.46&0.214&0.319&0.230&0.262&1.475&0.806\\
    \midrule
    \checkmark&\checkmark&\checkmark&\checkmark&$\times$&0.110&0.104&58.20&0.120&0.113&60.14&0.214&0.316&0.227&0.259&1.273&0.704\\
    \midrule
    \checkmark&\checkmark&\checkmark&\checkmark&\checkmark&0.111&0.105&58.38&0.120&0.113&60.27&0.212&0.317&0.225&0.259&1.078&0.671\\

    \bottomrule
    \end{tabular}}%
  \label{tab:ablation}%
\end{table*}%

\subsection{Main Results}

\noindent{\textbf{Long-term series forecasting.}}
Table~\ref{tab:main_result2} presents the comparison of the proposed approach with related methods on 6 long-term series forecasting datasets. 
Our method achieves the best or the second-best performance among existing methods across all datasets.
Notably, our method substantially outperforms the state-of-the-art on ETTm$_2$, Exchange, and ILI,
which are relatively small-scale datasets containing fewer than one million observation data points and a small number of variables per time step (typically 7 or 8).
{
Our method also outperforms the method of repeating the last values in the look-back windows according to the results as demonstrated in \cite{DBLP:conf/aaai/ZengCZ023}, which indicates that the mechanism of our method is unprecedented among all the existing financial models for the exchange rate prediction. 
We show that our method can actually capture information from the long-term observations rather than simply repeating or ``guessing'' the prediction from the latest observations in Subsection~\ref{subsec:computeeff}.
}
It validates that our method can handle more realistic scenarios, even with limited available information.
When going deeper into the results, we can find that the MAE scores are not always better than other methods on several datasets, where most MAE scores are lower than 0.5.
There are plausible explanations for this phenomenon.
One potential explanation is that our method excels at predicting relatively challenging samples (i.e., those with larger error scores exceeding 0.5) but slightly worse on some easily predictable samples (i.e., those with error scores below 0.5).
Alternatively, it might be attributed to the fact that there are fewer easy-to-predict samples with lower error scores than there are hard-to-predict samples with higher error scores. 
Achieving more precise predictions on the easy-to-predict samples is required to offset the impact of relatively inaccurate predictions on the hard-to-predict ones.
Our method's performance appears to align more with the former reason, as evidenced by its superior performance on ILI, which is rich in hard-to-predict samples, and its less ideal performance on relatively easy-to-predict datasets, i.e., ECL and Weather.

\noindent{\textbf{Stock trend forecasting.}}
Table~\ref{tab:main_result} compares the proposed method to commonly used methods for stock trend forecasting on CSI100 and CSI300.
As shown, our method outperforms the state-of-the-art on all benchmarks.
When combined with HIST, which exploits explicit and implicit enterprise information, our method (Ours+HIST) performs even better, suggesting that our method may be able to discover more latent factors for the time series or model the relations of the factors more efficiently.
Without using extra information, our method (Our+LR) also shows competitive results.
When compared with SFM which captures multi-frequency trading patterns inside the RNN cells, our method consistently outperforms it, highlighting the advantage of processing data within the latent concept domain.
Our method performs better than FactorVAE on most of the evaluation metrics, verifying its ability to discover more useful information from temporal variations.

\subsection{Ablation Studies}
\label{sub:ablation}

\begin{table*}[htbp]
  \scriptsize
  \centering
  \caption{Test results of different settings of signal components. The underlined set is selected for our method.}
    \setlength{\tabcolsep}{4mm}{
    \begin{tabular}{l|ccccccccc}
    \toprule
    \multicolumn{1}{l}{Sets of sampling rates}&\{1\}&\{1,2\}&\{1,3\}&\{1,5\}&\{1,2,3\}&\{\underline{1,2,5}\}&\{1,3,5\}&\{1,2,3,5\}&\{1,1,1\}\\
    \midrule
    IC&0.105&0.109&0.109&0.111&0.109&0.111&0.112&0.111&0.110\\
    Rank IC&0.100&0.103&0.103&0.104&0.102&0.105&0.105&0.106&0.104\\
    MP&58.26&58.32&58.27&58.41&58.32&58.38&58.34&58.36&58.31\\
    \bottomrule
    \end{tabular}}
  \label{tab:component}%
\end{table*}%

We conduct ablation experiments to study the effectiveness of different modules in our method.
The main modules of the proposed method are as follows:

\noindent\textbf{Decomposition (Decomp.)} represents that the input signal is decomposed into multiple signal components. The model without the decomposition denotes the input signal is just mapped into a single signal component.

\noindent\textbf{Disentanglement (Disent.)} represents the mechanism of using the KL divergence term for disentangling the inferred factors. The model without disentanglement is achieved by removing the KL divergence term from the objective function Eq.(\ref{eq:likelihood}) during training.

\noindent\textbf{Reconstruction (Recons.)} represents the objective of forcing the inferred factors to reconstruct the input signals at the next time step.  The model without the reconstruction is achieved by removing the reconstruction error from Eq. (\ref{eq:likelihood}) during training.

\noindent\textbf{Independence (Ind.)} represents that the factors in a signal component only depend on the previous factors in the same signal component. The model without independence is implemented by concatenating all the latent factors in different signal components to predict the next factors in a signal component.

\noindent{
\textbf{Shared Weights (S.W.)} represents that the weights are shared between the combinations of the sub-decoders and the predictors.
}

Table~\ref{tab:ablation} provides the experimental results of our ablation study conducted on CSI100, CSI300, Exchange, Weather, and ILI.
For the 3 long-term series forecasting datasets, we report the average MAE and MSE scores of the 4 prediction horizons.
Results in the 1-st and 2-nd rows reveal the importance of the decomposition module in enhancing task performance.
Without this model, the method achieves minimal improvement, indicating that the assumptions of predictability and independence play a crucial role in computational efficiency.
From the 3-rd row of the results, we can observe that disentanglement learning is helpful for improving performance.
It is probably because the semantics of the latent factors could be clarified for better representation with the disentangling operation.
Moreover, we also tried other disentanglement learning methods for handling the information bottleneck problem but got no improvement.
It might be because the reconstruction module helps to constrain the representation to be more informative, as supported by the findings in the 4-th row of the results. Without the guidance of the reconstruction module, learning performance tends to deteriorate.
{
The results of the 5-th row show that the performance would be slightly lower without sharing the attention weights because sharing weights provides guidance for mapping the reconstructions and predictions into the same space, leading to better convergence.
}
From the 6-th row of the results, we can safely conclude that there is no need to consider other signal components for the prediction of each signal component.

We investigate the settings of the signal components. 
We use different subsets $s\subseteq \{1,2,3,5\}$ of sampling rates for investigating the settings of the signal components on CSI100.
Table~\ref{tab:component} shows the comparison results.
We observe that models with sampling rate 5 have relatively good performance, because they explicitly present weekly signals (5 trading days).
Models with sampling rate 2 perform slightly better than those with sampling rate 3, and it might be because they can present monthly signals (20 trading days).
Factorizing signal components only using sampling rate 1 can theoretically present all the signals with different periodic trends, however, the performance degradation shows that explicit settings can improve the efficiency of the optimization.
The results indicate that the model would perform better with more diverse sampling rates, but we finally choose the setting $\{1,2,5\}$ for computational efficiency.
For the long-term series forecasting, we directly show the optimal sampling rates in Table~\ref{tab:samplingrate}.
Recall that the sampling rates are chosen from 1 and basic time units (i.e., hour, day, week, and month).
The sampling rate of 1 can be regarded as a residual operation.
For ETTm${_2}$, Exchange and ILI, all the time units are meaningful. 
Too large sampling rates are not chosen, because the sampled signals are too short to be modeled.
The hourly sampling is not used on the Weather dataset, which is rational because the changes of weather in the first minute and last minute in each hour do not contain any salient features.
Daily variants of different hours in ECL and Traffic are redundant for prediction according to the experiments.
It indicates that the daily scale contains no extra information than the hourly and weekly scales, and even introduces noise for the computation of the model.

\begin{table}[htbp]
\scriptsize
\centering
  \caption{Optimal sampling rates of different datasets.}
\setlength{\tabcolsep}{6mm}{
\begin{tabular}{lcc}
\toprule
Datasets  & Frequency & Sampling rates \\
\midrule
ETTm$_2$    & 15min     & \{1,4,96\}     \\
Exchange & 1day      & \{1,7,30\}     \\
ECL      & 1h        & \{1,168\}   \\
Traffic & 1h        & \{1,168\}   \\
Weather  & 10min     & \{1,144\}      \\
ILI      & 1week     & \{1,2,4\}     \\
\bottomrule
\end{tabular}}
\label{tab:samplingrate}%
\end{table}

We also conduct experiments of using different disentanglement operations for our model on CSI100, and the results are shown in Table~\ref{tab:disent}.
The hyper-parameters of the $\beta$-VAE$_{\textrm{B}}$ \cite{DBLP:journals/corr/abs-1804-03599} are set as follows.
The starting and final annealed capacities are 0 and 0.5, respectively.
The hyper-parameter, i.e., the coefficient of the TC term of the Factor-VAE \cite{DBLP:conf/icml/KimM18} is set to $0.6$.
Note that the Factor-VAE \cite{DBLP:conf/icml/KimM18} here is a totally different method from the FactorVAE \cite{duan2022factorvae}.
Although the $\beta$-VAE has the problem of information bottleneck when disentangling, and the  $\beta$-VAE$_{\textrm{B}}$  and the Factor-VAE aim to solve the problem, our method is still effective just using a simple disentanglement operation, i.e., increasing the value of $\beta$.
It might be because the sufficiency of our method can alleviate the bottleneck problem.
Although Factor-VAE yields the best performance, we still choose the simple strategy of increasing the $\beta$ for lower time complexity.

\begin{table*}[htbp]
  \scriptsize
  \centering
  \caption{Test results of different disentanglement operations. The underlined operation is selected for our method.}
    \setlength{\tabcolsep}{3mm}{
    \begin{tabular}{l|ccccc}
    \toprule
    \multicolumn{1}{l}{Disentanglement operations}&$\beta$-VAE ($\beta$=0.1) \cite{DBLP:conf/iclr/HigginsMPBGBML17}&$\beta$-VAE ($\beta$=1) \cite{DBLP:conf/iclr/HigginsMPBGBML17}&\underline{$\beta$-VAE (increasing $\beta$) \cite{DBLP:conf/iclr/HigginsMPBGBML17}}&$\beta$-VAE$_{\textrm{B}}$ \cite{DBLP:journals/corr/abs-1804-03599}&Factor-VAE \cite{DBLP:conf/icml/KimM18}\\
    \midrule
    IC&0.107&0.110&0.111&0.110&0.112\\
    Rank IC&0.102&0.104&0.105&0.105&0.106\\
    MP&58.43&58.32&58.38&58.28&58.39\\
    \bottomrule
    \end{tabular}}%
  \label{tab:disent}%
\end{table*}%

\subsection{Robustness to Insufficient Data}

\begin{figure}[tbp]
\centerline{\includegraphics[width=1\columnwidth]{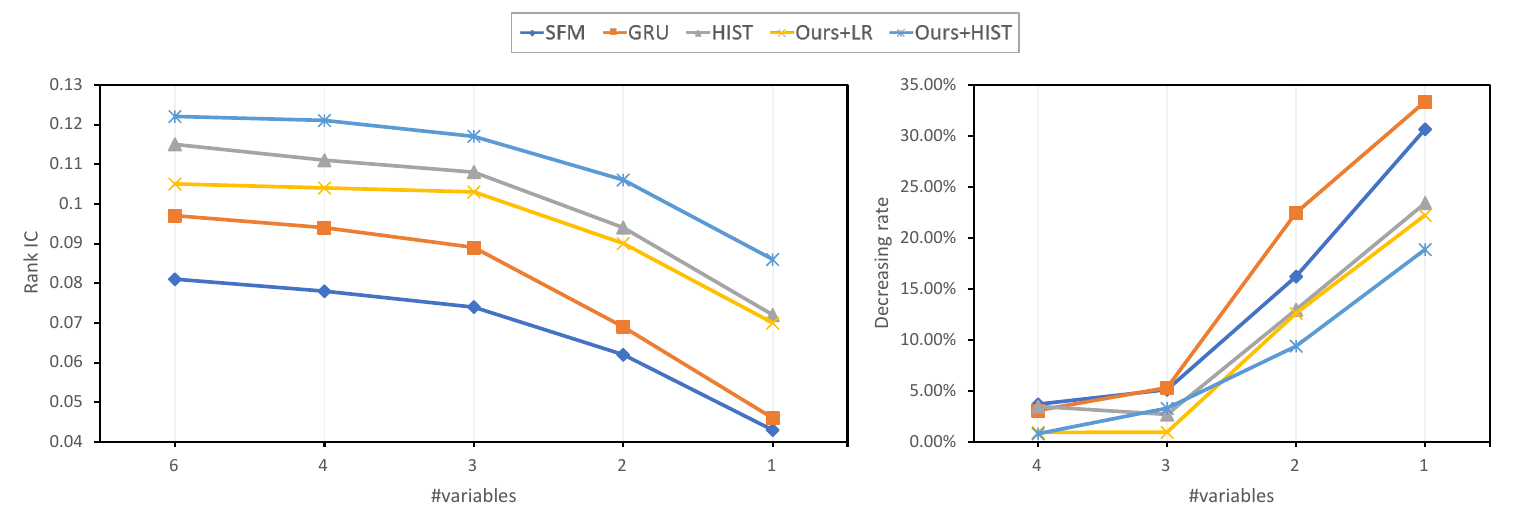}}
\caption{Performance degradation of different methods with insufficient input data on CSI100. The left panel shows the average rank IC scores of different methods with varying numbers of variables as input. The right panel shows the decreasing rates of these scores. }
\label{fig:insufficient}
\end{figure}

{
We investigate the efficacy of our method under conditions of insufficient observation data.
The CSI100 dataset for stock trend prediction is employed in this experiment, where the 6 variables of the input series signify distinct attributes: ``opening price'', ``closing price'', ``highest price'', ``lowest price'', ``volume weighted average price (VWAP)'' and ``trading volume''.
We conduct experiments by gradually removing variables from the input series.
The average rank IC score is computed upon removing certain numbers of different variables, along with the corresponding decreasing rate in rank IC.
Note that the variables of ``closing price'' and ``VWAP'' are excluded from our experiments, as the ``closing price'' can directly determine the prediction, and the VWAP annotations are extremely incomplete, with most values being 0.
As a result, we work with a reduced set of 4 candidate variables.
Fig.~\ref{fig:insufficient} presents the experimental results compared with other related methods.
The figure illustrates that our method declines slower than other methods, which is attributed to our method's capacity to fully discover latent factors from historical data.
While HIST also performs well, benefiting from the additional data, the result of Ours+LR affirms the heightened effectiveness of our mechanism in conditions of data insufficiency.
}

\begin{figure*}[tbp]
\centerline{\includegraphics[width=1.9\columnwidth]{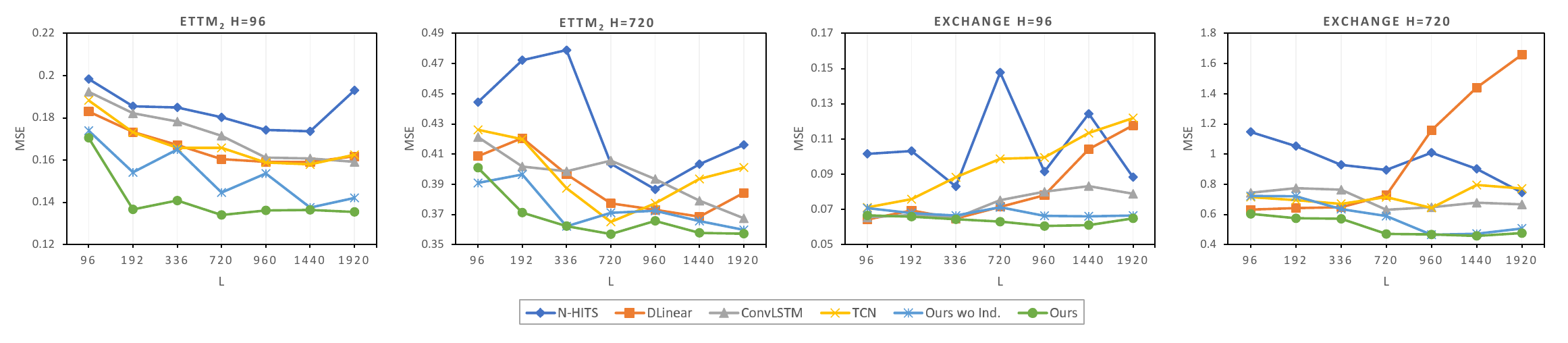}}
\caption{Test results (MSE) of prediction horizons $H=96,720$ with varying look-back windows $L=96,192,336,720,960,1440,1920$ on ETTm$_2$ and Exchange.}
\label{fig:long}
\end{figure*}

\subsection{Long-term Computational Efficiency}
\label{subsec:computeeff}

{
By leveraging the inference of predictable latent factors from observations and modeling sparse relations among them, our method excels at capturing long-term information from observed signals. 
To study the long-term computational efficiency of our method, we conduct a comparative analysis with several related techniques by evaluating the forecasting performances with progressively increasing look-back window size, $L$, as shown in Fig.~\ref{fig:long}.
Notably, when $L=96$ on ETTm$_2$, the sampling rate 96 is excluded as it makes no sense for an RNN with only 1 time step input. 
In addition to benchmarking against 2 state-of-the-art methods, namely N-HITS and DLinear (both implemented using the authors' provided source codes), we evaluated 3 additional methods to explore specific mechanisms for handling long-term series of our method.
These include our backbone model - a fusion of 1D convolutional layers and LSTMs (ConvLSTM), cascaded 1D dilated convolutional layers employing the same dilation rates as our method (TCN), and our model without the independence operation (ours wo Ind.).
}

{
The figure shows that our methods can constantly extract useful information from observations.
This is evident from the generally ascending trend in our method's performances because our method's performance exhibits a generally ascending trend with the look-back window size increasing on both datasets.
For the Exchange dataset, the lowest MSE scores are found between the look-back window sizes of 720 and 960. 
This trend likely arises from the model's limited capability to process extensive information on the Exchange dataset. 
However, it is more likely that longer series no longer contain additional useful information for forecasting, as evidenced by the majority of optimal results of other methods being observed at $L\leq 336$.
The trend pattern of ours wo Ind. is similar to our method, suggesting the overall framework of our method facilitates the extraction of long-term information, and the independent operation further improves the efficacy.
ConvLSTM shows its effectiveness in long-term computation, primarily owing to the long-term memory mechanism of LSTM. 
We believe that our method also benefits from this aspect.
Conversely, cascaded multi-scale sampling without decomposition contributes minimally to long-term computational efficiency improvement, as indicated by TCN's outcomes.
In conclusion, the long-term computational efficiency of our method stems primarily from the independence operation, the long-term memory mechanism, and the parallel multi-scale sampling with decomposition.
}

\begin{figure}[tbp]
\centerline{\includegraphics[width=0.93\columnwidth]{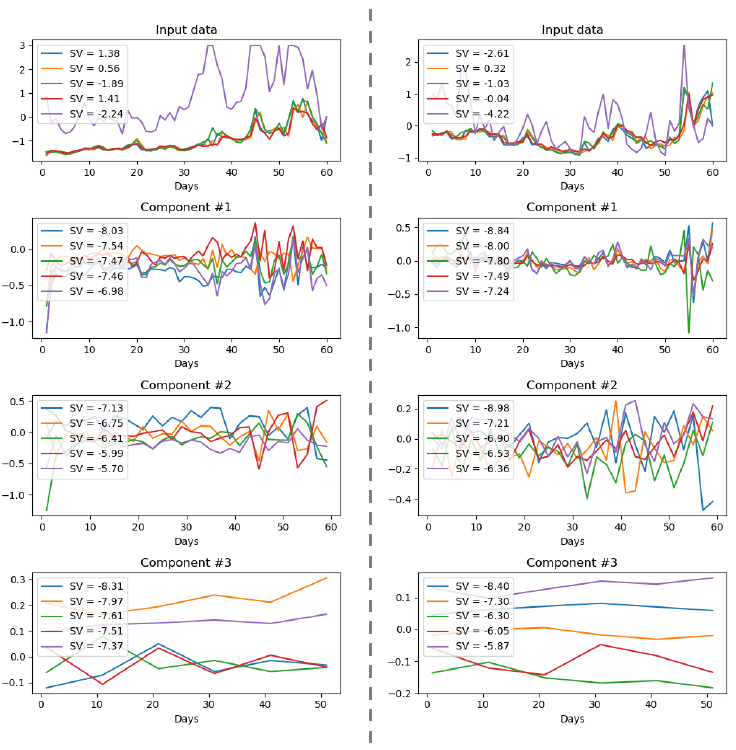}}
\caption{Examples of the ADF tests on the original signal and learned factors. Each column presents the visualization results of a randomly selected sample.}
\label{fig:sa}
\end{figure}

\begin{figure}[htbp]
\centerline{\includegraphics[width=1\columnwidth]{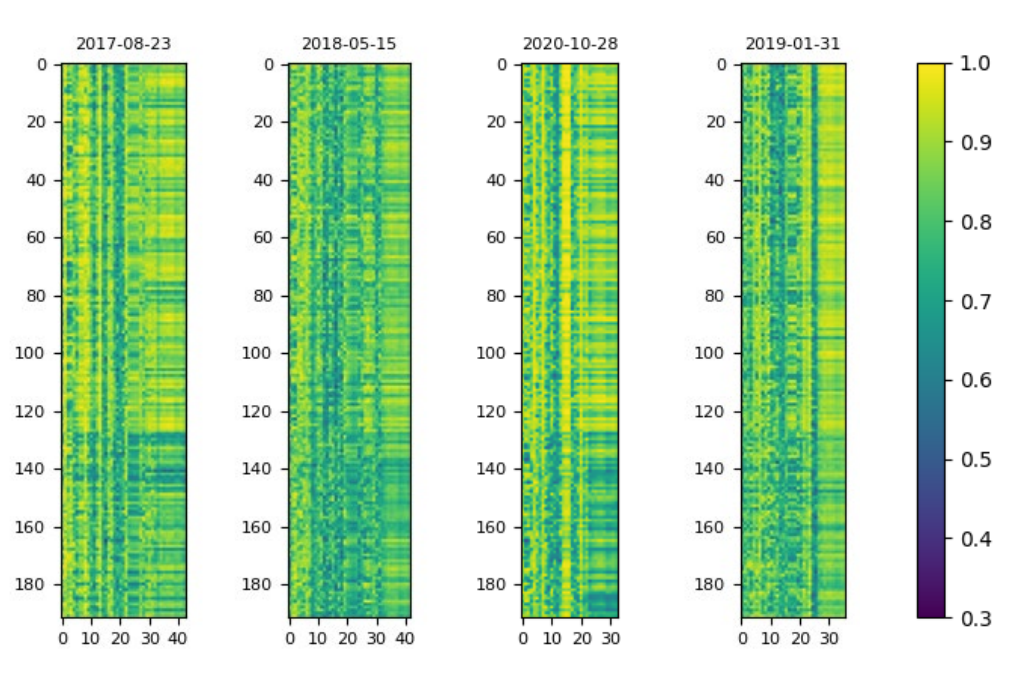}}
\caption{
Visualization of correspondences between latent factors and predefined concepts in the stock market on CSI100 on 4 different days. 
The x-axis represents concept indices and the y-axis lists latent factor indices.
The correspondence is measured by the ROC-AUC score calculated by classifying stocks into a certain concept according to the clustering results of the time series of a latent factor. 
A higher correspondence value indicates that the latent factor is more likely to share some meaningful properties with an explicit concept.}
\label{fig:concept}
\end{figure}

\subsection{Statistical Analysis}

The predictability of the latent factor is encouraged in our model.
According to Eq.(\ref{eq:predicability}), stationarity is a necessary condition of a predictable time series, since stationarity ensures that the function $g$ is able to output similar values for similar inputs.
It's difficult to design appropriate statistics for testing predictability, but lots of hypothesis test methods for stationarity are proposed.
Here, the classical Augmented Dickey-Fuller unit root test (ADF test) is used on both the input original price signal and the learned latent factors, and Fig. \ref{fig:sa} depicts the test results of 2 randomly selected samples.
The first row shows the original price paths along time (we neglect the VWAP, because the values are almost zero).
The other rows show the signal components with each color representing a representative latent factor with the topmost statistic values (SVs).
As expected, the SVs of the learned factors are commonly less than the critical value under the significance level of $1\%$, $5\%$ or $10\%$, i.e., $-3.46$, $-2.87$, $-2.57$, indicating that the latent factor series is stationary.
The ADF tests the linear reliability contained in the time series, and it is interesting to note that our factors can still refuse the null hypothesis even though the $g$ in the proposed method is implemented via GRU which is a non-linear neural network.
The probable reason is that the learned $g$ actually approximates an injective to fit the complex data for predictability, which is more strict than the stationarity.

{
To study the interpretability of the inferred latent factors, we draw heatmaps illustrating the correspondence between latent factors and predefined concepts within the stock market, as depicted in Fig.~\ref{fig:concept}. 
This exploration aims to determine whether the latent factors bear relevance to tangible real-world semantics.
Thankfully, the Alpha360 dataset provides annotations to predefined concepts in the stock market, such as technology and E-Commerce, and these annotations are reprocessed as a binary matrix with each element assuming a value of 0 or 1, denoting whether a stock belongs to be a given concept by \cite{DBLP:journals/corr/abs-2110-13716}, and used in their HIST framework as extra knowledge.
We postulate that stocks categorized under the same predefined stock concept share analogous inherent patterns.
Some of these patterns may align with our learned latent factors. 
If a simple unsupervised classifier can accurately recognize whether the stocks belong to a concept with one of the latent factors, it implies that the latent factor represents meaningful characteristics related to the concept.
If the concept classification results derived from a latent factor are constantly the same over time, it would further validate that our model can learn the latent concept with a particular semantic.
Accordingly, we use $k$-means clustering to classify the stocks with the 60-dimensional historical data of each latent factor as features.
For each concept and each latent factor, we compute the ROC-AUC scores for all the clusters.
The highest of these scores is selected as a quantification of the correspondence between the given concept and the latent factor.
In this experiment, we set the number of latent factors in each signal component as 64 for clear visualization.
Note that we remove concepts that contain less than 3 stocks, and because the stocks vary over time, the concepts of each day would be slightly different.
And the order of the concepts remains unchanged.
From the figure, many latent factors show high correspondence between the stock concepts, and the whole pattern of each day is generally the same, demonstrating that the latent factors are very likely to have meaningful semantics.
}

\section{Conclusions}

We have introduced a simple but efficient method of inferring predictable latent factors for time series forecasting.
The inferred factors can form multiple independent signal components with characteristics of predictability, sufficiency, and identifiability to guarantee that our model effectively reconstructs the future values with these components in the latent concept domain.
Thanks to the independence and predictability of the discovered signal components, our method enables sparse relation modeling of the inferred factors for long-term efficiency and the easy reconstruction of the prediction.
Experiments validate the effectiveness of the proposed method and the predictability of the inferred latent factors.

The limitation of our work is that we empirically sample data of different time scales for inferring the latent factors and the signal components with specific inductive bias.
In the future, we plan to explore a more general method that enables the model to learn the optimal encoding structure by adding a meta-learning algorithm.

\section*{Acknowledgement}
This work was supported in part by the National Key Research and Development Program of China under grant No. 2020YFC1523200, and the Natural Science Foundation of China under grants No. 62106021 and No. U20A20225.

\bibliographystyle{IEEEtran}
\bibliography{ref}

\end{document}